\documentclass[12pt]{article}

\usepackage{latexsym}
\usepackage{psfrag}

\input{epsf}

\title{Tabular Parsing}
\author{
\begin{tabular}[t]{c}
Mark-Jan Nederhof%
  \,\thanks{
Supported by the Royal Netherlands
Academy of Arts and Sciences.
Secondary affiliation is the
German Research Center for Artificial Intelligence (DFKI).
} \\
Faculty of Arts \\
University of Groningen \\
P.O.\ Box 716 \\
NL-9700 AS Groningen, The Netherlands \\
{\tt markjan@let.rug.nl}
\end{tabular}
\and
\begin{tabular}[t]{c}
Giorgio Satta \\
Department of Information Engineering \\
University of Padua \\
via Gradenigo, 6/A \\
I-35131 Padova, Italy \\
{\tt satta@dei.unipd.it}
\end{tabular}
}
\date{}

\newcommand{\size}[1]{\left | {#1} \right |}
\newcommand{\order}[1]{{\cal O}({#1})}

\newcommand{\sep}{\,\mid\,}

\newcommand{\mygram}{{\cal G}}
\newcommand{\de}{\rightarrow}
\newcommand{\desingle}{\Rightarrow}
\newcommand{\dem}{\Rightarrow^\ast}

\newcommand{\myaut}{{\cal A}}

\newcommand{\myterm}{{\mit\Sigma}}
\newcommand{\mynont}{N}
\newcommand{\myrule}{R}

\newcommand{\bul}{\mathrel{\bullet}}

\newcommand{\ep}{\varepsilon}

\newcommand{\mysym}{Q}
\newcommand{\Xinit}{q_{\it init}}
\newcommand{\Xfinal}{q_{\it final}}
\newcommand{\mytransset}{\mit\Delta}

\newcommand{\mytrans}[3]{{#1} \stackrel{#2}{\mapsto} {#3}}

\newcommand{\mymove}{\vdash}
\newcommand{\mymoves}{\vdash^\ast}

\newcommand{\mytable}{{\cal T}}
\newcommand{\mynew}{{\cal N}}

\newcommand{\mygoto}{{\it goto\/}}
\newcommand{\myclosure}{{\it closure\/}}

\newcommand{\tabrulestart}[2]{
$ 
        \frac{ \begin{array}{c} #1 \end{array} }
                        { \begin{array}{c} #2 \end{array} }
$} 

\newcommand{\tabrulestartlabel}[3]{
\begin{eqnarray}
\label{#1}
        \frac{ \begin{array}{c} #2 \end{array} }
                        { \begin{array}{c} #3 \end{array} }
\end{eqnarray} 
}

\newcommand{\tabrule}[3]{
$ 
        \frac{ \begin{array}{c} #1 \end{array} }
                        { \begin{array}{c} #2 \end{array} }
   \left\{ \begin{array}{l} #3 \end{array} \right.
$}

\newcommand{\tabrulelabel}[4]{
\begin{eqnarray}
  \label{#1}
        \frac{ \begin{array}{c} #2 \end{array} }
                        { \begin{array}{c} #3 \end{array} }
   \left\{ \begin{array}{l} #4 \end{array} \right.
  \end{eqnarray}
}

\psfrag{SbulS+S}{$S\de\; \bul S+S$}
\psfrag{SSbul+S}{$S\de  S\bul+S$}
\psfrag{SS+bulS}{$S\de  S+ \bul S$}
\psfrag{SS+Sbul}{$S\de  S+S\bul$}
\psfrag{Sbula}{$S\de\; \bul a$}
\psfrag{Sabul}{$S\de   a\bul$}

\psfrag{S}{$S$}
\psfrag{a}{$a$}
\psfrag{+}{$+$}

\psfrag{qinit}{$\Xinit$}
\psfrag{qfinal}{$\Xfinal$}
\psfrag{bot}{$\bot$}
\psfrag{q0}{$q_0$}
\psfrag{q1}{$q_1$}
\psfrag{q2}{$q_2$}
\psfrag{q3}{$q_3$}
\psfrag{q4}{$q_4$}
\psfrag{q5}{$q_5$}
\psfrag{q6}{$q_6$}
\psfrag{q7}{$q_7$}
\psfrag{q8}{$q_8$}
\psfrag{q9}{$q_9$}

\begin{document}
\maketitle

\section{Introduction}
\label{NedSat:intro}

{\em Parsing\/} is the process of determining the parses of an
input string according to a grammar. 
In this chapter
we will restrict ourselves to context-free grammars.
Parsing is related to {\em recognition\/}, 
which is the process of determining
whether an input string is in the language described by a grammar
or automaton.
Most algorithms we will discuss are recognition algorithms,
but since they can be straightforwardly extended 
to perform parsing, we will not make a sharp distinction here
between parsing and recognition algorithms.

For a given grammar and an input string,
there may be very many parses, 
perhaps too many to be enumerated one by one. Significant practical
difficulties in computing and storing the parses can be
avoided by computing individual fragments of these
parses and storing them in a table. The advantage of this is
that one such fragment may be shared by many different parses.
The methods of {\em tabular parsing\/} that we will investigate
in this chapter are capable of computing and representing
exponentially many parses in polynomial time and space, respectively,
by means of this idea of sharing of fragments between
several parses.

Tabular parsing, invented in the field of computer science
in the period roughly between 1965 and 1975, also became known
later in the field of computational linguistics as
{\em chart parsing\/}~\cite{TH84}. 
Tabular parsing is a form of {\em dynamic programming\/}.
A very related approach is to apply {\em memoization\/}
to functional parsing algorithms~\cite{LE93}.

What is often overlooked in modern parsing literature is
that many techniques of tabular parsing can be straightforwardly
derived from non-tabular parsing techniques expressed by means of
push-down automata. A push-down automaton is a device that
reads input from left to right, while manipulating a 
stack. Stacks are a very common data structure, 
frequently used wherever there is recursion, 
such as for the implementation of functions and procedures 
in programming languages, but also for context-free parsing.

Taking push-down automata as our starting point 
has several advantages for describing tabular parsers. 
Push-down automata are simpler devices
than the tabular parsers that can be derived from them.
This allows us to get acquainted with simple,
non-tabular forms of context-free parsing before we move
on to tabulation, which can, to a large extent, be explained independently
from the workings of individual push-down automata. Thereby we 
achieve a separation of concerns.
Apart from these presentational advantages,
parsers can also be implemented more easily 
with this modular design than without.

In Section~\ref{NedSat:pda} we discuss push-down
automata and their relation to context-free grammars.
Tabulation in general is introduced in Section~\ref{NedSat:tabulation}.
We then discuss a small number of specific tabular parsing algorithms
that are well-known in the literature, viz.\ 
Earley's algorithm (Section~\ref{NedSat:Earley}),
the Cocke-Kasami-Younger algorithm (Section~\ref{NedSat:CKY}),
and tabular LR parsing (Section~\ref{NedSat:GLR}).
Section~\ref{NedSat:forests} discusses compact representations of
sets of parse trees, which can be computed by tabular parsing algorithms.
Section~\ref{NedSat:literature} provides further pointers
to relevant literature.

\section{Push-down automata}
\label{NedSat:pda}

The notion of push-down automaton plays a central role in this chapter.
Contrary to what we find in some textbooks, our push-down automata do not
possess states next to stack symbols. This is without loss
of generality, since states can be encoded into the stack symbols.
Thus,
a {\em push-down automaton\/} (PDA) $\myaut$ is a 5-tuple
$(\myterm,$ $\mysym,$ $\Xinit,$ $\Xfinal,$ $\mytransset)$,
where $\myterm$ is an alphabet,
i.e., a finite set of input symbols,
$\mysym$ is a finite set of {\em stack symbols},
including the {\em initial stack symbol\/} $\Xinit$ and the
{\em final stack symbol\/} $\Xfinal$, and $\mytransset$ is 
a finite set of {\em transitions}.

A transition has the form
$\mytrans{\sigma_1}{v}{\sigma_2}$, where
$\sigma_1,\sigma_2 \in \mysym^\ast$ and
$v \in \myterm^\ast$.
Such a transition can be applied if the stack symbols
$\sigma_1$ are found to be
the top-most few symbols on the stack and the input symbols $v$ 
are the first few symbols of the unread part of the input.
After application of such a transition, $\sigma_1$ has
been replaced by $\sigma_2$, and the next $\size{v}$ input symbols
are henceforth treated as having been read.

More precisely, for a fixed PDA and 
a fixed input string $w= a_1 \cdots a_n \in \myterm^\ast$,
$n \geq 0$, 
we define a {\em configuration\/} as
a pair $(\sigma,i)$ consisting of a stack $\sigma\in \mysym^\ast$ and
an input position $i$, $0 \leq i \leq n$.
The {\em input position\/} indicates how many of the symbols from the input
have already been read.  Thereby, position 0 and position $n$ 
indicate the beginning and the end, respectively, of $w$. 
We define the binary relation
$\mymove$ on configurations by:
$(\sigma,i)\mymove (\sigma',j)$ if and only
if there is some transition
$\mytrans{\sigma_1}{v}{\sigma_2}$
such that
$\sigma = \sigma_3 \sigma_1$ and
$\sigma' = \sigma_3 \sigma_2$, some $\sigma_3 \in \mysym^\ast$,
and $v=a_{i+1} a_{i+2} \cdots a_{j}$.
Here we assume $i \leq j$, and if $i=j$ then
$v=\ep$, where $\ep$ denotes the empty string.
Note that
in our notation, stacks grow from left to right, i.e., the top-most
stack symbol will be found at the right end.

We denote the reflexive and transitive closure
of $\mymove$ by $\mymoves$; in other words,
$(\sigma,i)$ $\mymoves$ $(\sigma',j)$ means that
we may obtain
configuration $(\sigma',j)$ from $(\sigma,i)$ by applying
zero or more transitions. We say that
the PDA {\em recognizes\/} a 
string $w= a_1 \cdots a_n$ if
$(\Xinit,0)$ $\mymoves$ $(\Xfinal,n)$. This means that
we start with a stack containing only the initial
stack symbol, and the input position is initially 0,
and recognition is achieved if we succeed in reading
the complete input, up to the last position $n$, while
the stack contains only the final stack symbol.
The language {\em accepted\/} by a PDA is the
set of all strings that it recognizes.

As an example, consider the PDA with
$\myterm = \{a,b,c,d\}$,
$\mysym = \{q_0, \ldots, q_9\}$,
$\Xinit = q_0$,
$\Xfinal = q_9$,
and the set $\mytransset$ of transitions given in
Figure~\ref{NedSat:exPDA}.
\begin{figure}[t]
\hspace*{\fill}
\begin{minipage}[t]{2cm}
$$
\begin{array}[t]{l}
\mytrans{q_0}{a}{q_0\ q_1} \\
\mytrans{q_0\ q_1}{b}{q_0\ q_2} \\
\mytrans{q_0\ q_1}{b}{q_0\ q_3} 
\end{array}
$$
\end{minipage}
\hfill
\begin{minipage}[t]{2cm}
$$
\begin{array}[t]{l}
\mytrans{q_2}{c}{q_2\ q_4} \\
\mytrans{q_3}{c}{q_3\ q_4}  \\
\mytrans{q_4}{d}{q_4\ q_5} 
\end{array}
$$
\end{minipage}
\hfill
\begin{minipage}[t]{2cm}
$$
\begin{array}[t]{l}
\mytrans{q_4\ q_5}{\ep}{q_6} \\
\mytrans{q_2\ q_6}{\ep}{q_7} \\
\mytrans{q_0\ q_7}{\ep}{q_9} 
\end{array}
$$
\end{minipage}
\hfill
\begin{minipage}[t]{2cm}
$$
\begin{array}[t]{l}
\mytrans{q_3\ q_6}{\ep}{q_8} \\
\mytrans{q_0\ q_8}{\ep}{q_9} 
\end{array}
$$
\end{minipage}
\hspace*{\fill}
\caption{Transitions of an example PDA.}
\label{NedSat:exPDA}
\end{figure}
There are two ways of recognizing
the input string $w=a_1a_2a_3a_4={\it abcd}$, indicated by the two
sequences of configurations in Figure~\ref{NedSat:2recognized}.
\begin{figure}[t]
\hspace*{\fill}
\begin{minipage}{4cm}
$$
\begin{array}{|l|l|}
\hline
q_0 & 0 \\
q_0\ q_1 & 1 \\
q_0\ q_2 & 2 \\
q_0\ q_2\ q_4 & 3 \\
q_0\ q_2\ q_4\ q_5 & 4 \\
q_0\ q_2\ q_6 & 4 \\
q_0\ q_7 & 4 \\
q_9 & 4 \\
\hline
\end{array}
$$
\end{minipage}
\hfill
\begin{minipage}{4cm}
$$
\begin{array}{|l|l|}
\hline
q_0 & 0 \\
q_0\ q_1 & 1 \\
q_0\ q_3 & 2 \\
q_0\ q_3\ q_4 & 3 \\
q_0\ q_3\ q_4\ q_5 & 4 \\
q_0\ q_3\ q_6 & 4 \\
q_0\ q_8 & 4 \\
q_9 & 4 \\
\hline
\end{array}
$$
\end{minipage}
\hspace*{\fill}
\caption{Two sequences of configurations, leading to 
recognition of the string ${\it abcd}$.}
\label{NedSat:2recognized}
\end{figure}

We say a PDA is {\em deterministic\/} if for each
configuration there can be at most one applicable
transition.
The example PDA above is clearly {\em nondeterministic\/}
due to the two transitions 
$\mytrans{q_0\ q_1}{b}{q_0\ q_2}$ and
$\mytrans{q_0\ q_1}{b}{q_0\ q_3}$.

A {\em context-free grammar\/} (CFG) $\mygram$ is 
a 4-tuple $(\myterm, \mynont, S, \myrule)$,
where $\myterm$ is an {\em alphabet}, i.e., a finite set of {\em terminals}, 
$\mynont$ is a finite set of {\em nonterminals},
including the {\em start symbol\/} $S$, and $\myrule$ is a finite set of
{\em rules}, each of the form $A\de\alpha$ with $A\in \mynont$ and
$\alpha\in (\myterm \cup \mynont)^\ast$.
The usual `derives' relation is denoted by
$\desingle$, and its reflexive and transitive
closure by $\dem$. The language {\em generated\/} by a CFG
is the set $\{w \sep S \dem w\}$.

In practice, a PDA is not hand-written, but
is automatically obtained from a CFG, by a mapping
that preserves the generated/accepted language.
Particular mappings from CFGs to PDAs can be
seen as formalizations of {\em parsing strategies\/}. 

We define the size of a PDA as 
$\sum_{(\mytrans{\sigma_1}{v}{\sigma_2}) \in \mytransset} 
\size{\sigma_1 v \sigma_2}$, i.e., the total number of occurrences of
stack symbols and input symbols in the set of transitions.
Similarly, we define the size of a CFG as 
$\sum_{(A \de \alpha) \in \myrule} \size{A\alpha}$,
i.e., the total number of occurrences of
terminals and nonterminals in the set of rules.

\section{Tabulation}
\label{NedSat:tabulation}

In this section, we will restrict the allowable transitions
to those of the types
$\mytrans{q_1}{a}{q_1\ q_2}$,
$\mytrans{q_1\ q_2}{a}{q_1\ q_3}$,
and $\mytrans{q_1\ q_2}{\ep}{q_3}$,
where $q_1,q_2,q_3 \in \mysym$ and $a \in \myterm$.
The reason is that this allows a very simple form of tabulation,
based on work by~\cite{LA74,BI89}.
In later sections, we will again consider less restrictive types
of transitions. Note that each of the transitions in Figure~\ref{NedSat:exPDA}
is of one of the three types above.

The two sequences of configurations in Figure~\ref{NedSat:2recognized}
share a common step, viz.\ the application of transition
$\mytrans{q_4}{d}{q_4\ q_5}$ at input position $3$ when
the top-of-stack is $q_4$. In this section we will show how we
can avoid doing this step twice. Although the savings in time and
space for this toy example are negligible, in realistic examples
we can reduce the costs from exponential to polynomial, as we 
will see later.

A central observation is that if two configurations share the
same top-of-stack and the same input position, then the sequences
of steps we can perform on them are identical as long as we do not
access lower regions of the stack that differ between these
two configurations. This implies for example that in order to determine
which transition(s) of the form $\mytrans{q_1}{a}{q_1\ q_2}$
to apply, we only need to know 
the top-of-stack $q_1$, and the current input position 
so that we can check whether $a$ is the next unread symbol
from the input.

These considerations lead us to propose a representation of
sets of configurations as graphs. 
The set of vertices is partitioned
into subsets, one for each input position, 
and each such subset contains at most one vertex for each 
stack symbol. This last condition
is what will allow us to share steps
between different configurations.

We also need arcs in the graph to connect the stack symbols.
This is necessary when transitions of the form
$\mytrans{q_1\ q_2}{a}{q_1\ q_3}$
or $\mytrans{q_1\ q_2}{\ep}{q_3}$ are applied,
since these require access to deeper regions of a stack
than just its top symbol. The graph will contain
an arc from a vertex representing stack symbol $q$ at position $i$
to a vertex representing stack symbol $q'$ at position $j \leq i$, 
if $q'$ resulted as the topmost stack symbol at input
position $j$, and $q$ can be immediately on top of that $q'$
at position $i$. If we take a path from a vertex in the
subset of vertices for position $i$, and follow arcs 
until we cannot go any further,
encountering stack symbols $q_1$, \ldots, $q_m$, in this order, 
then this means that $(\Xinit,0)$ $\mymoves$ $(q_m\cdots q_1,i)$.

\begin{figure}[t]
\begin{center}
\includegraphics{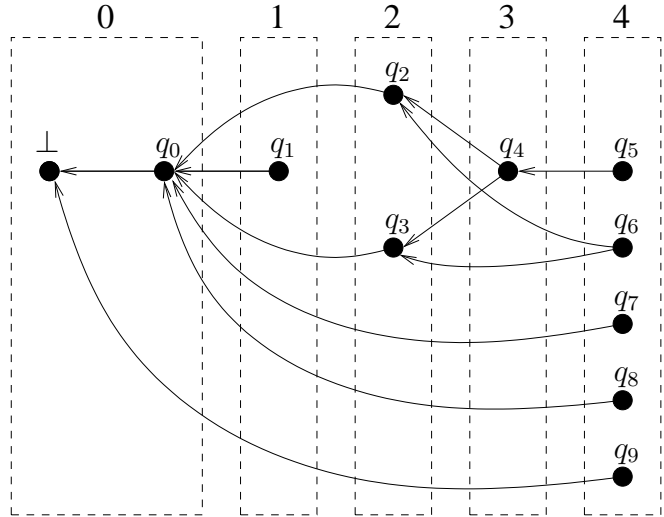}
\end{center}
\caption{The collection of all derivable configurations 
represented as graph. 
For each input position there is a subset of 
vertices. For each such subset, there is at most
one vertex for each stack symbol.}
\label{NedSat:graph}
\end{figure}

For the running example, the graph after completion of
the parsing process is given in Figure~\ref{NedSat:graph}. 
One detail we have not yet mentioned is that we need an
imaginary stack symbol $\bot$, which we assume
occurs below the actual bottom-of-stack. We need
this symbol to represent stacks consisting of a single
symbol. Note that the path from the vertex labelled
$q_9$ in the subset for
position 4 to the vertex labelled $\bot$ means that 
$(q_0,0)$ $\mymoves$ $(q_9,4)$, which implies
the input is recognized.

What we still need to explain is how we can construct the
graph, for a given PDA and input string. 
Let $w = a_1 \cdots a_n$, $n \geq 0$, be an input string.
In the algorithm
that follows, we will manipulate 4-tuples $(j,q',i,q)$, 
where $q',q\in\mysym$ and $j, i$ are 
input positions with $0 \leq j \leq i \leq n$. 
These 4-tuples will be called {\em items\/}.   Item $(j,q',i,q)$
means that there is an arc in the graph from a vertex representing
$q$ at position
$i$ to a vertex representing $q'$ at position $j$. Formally, it means that
for some $\sigma$ we have
$(\Xinit,0) \mymoves (\sigma\ q',j)$ and  
$(\sigma\ q',j) \mymoves (\sigma\ q'\ q, i)$, where in the 
latter relation the transitions that are involved do not access any symbols
internal to $\sigma$.

The algorithm is given in Figure~\ref{NedSat:generic_alg}.
Initially,
we let the set $\mytable$ contain only the item 
$(\bot,0,\Xinit,0)$, representing one arc in the graph.
We then incrementally fill $\mytable$ with more items, representing
more arcs in the graph, until the complete graph has been constructed.
In this particular tabular algorithm, we process
the symbols from the input one by one, from left to right, 
applying all transitions
as far as we can before moving on to the next input symbol.
Whereas $\mytable$ contains all items that have been derived up to
a certain point, the set $\mynew$ contains only those
items from $\mytable$ that still need to be
combined with others in order to (possibly) obtain new items. 
The set 
$\mytable$ will henceforth be called the {\em table\/} and
the set $\mynew$ the {\em agenda\/}.
\begin{figure}[t]
\begin{enumerate}
\item Let $\mytable$ $=$ $\{(\bot,0,\Xinit,0)\}$.
\item For $i = 1, \ldots, n$ do:
\begin{enumerate}
\item Let $\mynew$ $=$ $\emptyset$.
\item For each $(q',j,q_1,i-1)\in \mytable$ and each transition
$\mytrans{q_1}{a_i}{q_1\ q_2}$ such that 
$(q_1,i-1,q_2,i)\notin\mytable$, 
add $(q_1,i-1,q_2,i)$ to $\mytable$ and to $\mynew$.
\item For each $(q_1,j,q_2,i-1)\in \mytable$
and each transition
$\mytrans{q_1\ q_2}{a_i}{q_1\ q_3}$ such that
$(q_1,j,q_3,i)\notin\mytable$,
add $(q_1,j,q_3,i)$ to $\mytable$ and to $\mynew$.
\item As long as $\mynew \neq \emptyset$ do:
\begin{enumerate}
\item Remove some $(q_1,j,q_2,i)$ from $\mynew$.
\item For each $(q',k,q_1,j)\in \mytable$ and each transition
$\mytrans{q_1\ q_2}{\ep}{q_3}$ and
$(q',k,q_3,i)\notin\mytable$,
add $(q',k,q_3,i)$ to $\mytable$ and to $\mynew$.
\end{enumerate}
\end{enumerate}
\item Recognize the input if $(\bot,0,\Xfinal,n)\in\mytable$.
\end{enumerate}
\caption{Tabular algorithm to find the collection of all derivable configurations
for input $a_1\cdots a_n$,
in the form of a set $\mytable$ of items.}
\label{NedSat:generic_alg}
\end{figure}

Let us analyze the worst-case time complexity of 
the algorithm in Figure~\ref{NedSat:generic_alg}. 
We assume that the table $\mytable$ is implemented as a
square array of size $n+1$, indexed by input positions $i$ and $j$,
and that each item can be stored in and retrieved from $\mytable$ in 
time $\order{1}$.  The agenda $\mynew$ can be implemented 
as a stack. 
Let us consider Step~2(d).  A single application of this
step takes time $\order{1}$.  Since each such application is uniquely 
identified by a transition $\mytrans{q_1\ q_2}{\ep}{q_3}$,
a stack symbol $q'$ and the three input positions $i$, $j$ and $k$, 
the number of possible applications of the step is 
$\order{\size{\mytransset} \size{\mysym} n^3 }$, 
which for our PDAs can be rewritten as
$\order{\size{\myaut} \size{\mysym} n^3 }$. 
It is not difficult to see that this quantity
also dominates the worst-case time complexity of our algorithm, 
which is thereby polynomial both in the size of the 
PDA and in the length of the input string. 
A similar analysis shows that the space complexity 
of the algorithm is $\order{\size{\mysym}^2 n^2 }$. 

Although the use of the agenda in the algorithm 
from Figure~\ref{NedSat:generic_alg} allows a fairly
straightforward implementation, it obscures somewhat how items are
derived from other items. This can be described more clearly
by abstracting away from certain details of the algorithm,
such as the order in which items are added to $\mytable$.
This can be achieved by means of a
{\em deduction system\/}~\cite{SH95}.\footnote{The earliest mention
of abstract specifications of parsing algorithms may be due to~\cite{CO70a}. 
See also~\cite{SI97}.} 
Such a system contains a set of inference rules, each 
consisting of a list of {\em antecedents\/}, which stand
for items that we have already established to be in $\mytable$,
and, below a horizontal line, the {\em consequent\/},
which stands for an item that we derive from the antecedents
and that is added to $\mytable$ unless it is already present.
At the right of an inference rule, we may also write a number
of {\em side conditions\/}, which indicate when 
rules may be applied, on the basis of transitions of the PDA.

A deduction system equivalent to the algorithm from Figure~\ref{NedSat:generic_alg}
is given in Figure~\ref{NedSat:deduction}.
\begin{figure}[t]
\hspace*{\fill}
\begin{minipage}{6.5cm}
\tabrulestart{ \
}{
(\bot,0,\Xinit,0)
}
\\[2ex]

\tabrule{ 
(q',j,q_1,i-1)
}{
(q_1,i-1,q_2,i)
}{
\mytrans{q_1}{a_i}{q_1\ q_2}
}
\end{minipage}
\hfill
\begin{minipage}{6.5cm}
\tabrule{ 
(q_1,j,q_2,i-1)
}{
(q_1,j,q_3,i)
}{
\mytrans{q_1\ q_2}{a_i}{q_1\ q_3}
}
\\[2ex]

\tabrule{ 
(q',k,q_1,j) \\
(q_1,j,q_2,i)
}{
(q',k,q_3,i)
}{
\mytrans{q_1\ q_2}{\ep}{q_3}
}
\end{minipage}
\hspace*{\fill}

\caption{Tabular parsing algorithm in the form of a
deduction system.}
\label{NedSat:deduction}
\end{figure}
In Figure~\ref{NedSat:graph},
$(0,q_0,1,q_1)$ is derived from
$(\bot,0,q_0,0)$ by means of $\mytrans{q_0}{a}{q_0\ q_1}$,
$a$ being $a_1$;
$(0,q_0,2,q_2)$ is derived from
$(0,q_0,1,q_1)$ by means of $\mytrans{q_0\ q_1}{b}{q_0\ q_2}$,
$b$ being $a_2$;
$(0,q_0,4,q_7)$ is derived from
$(0,q_0,2,q_2)$ and $(2,q_2,4,q_6)$ 
by means of $\mytrans{q_2\ q_6}{\ep}{q_7}$.

We may now extend our repertoire of transitions by
those of the forms
$\mytrans{q_1}{\ep}{q_1\ q_2}$ and
$\mytrans{q_1\ q_2}{\ep}{q_1\ q_3}$,
which only requires two additional inference rules, 
indicated in Figure~\ref{NedSat:morerules}.
\begin{figure}[t]
\hspace*{\fill}
\begin{minipage}{6.5cm}
\tabrule{
(q',j,q_1,i)
}{
(q_1,i,q_2,i)
}{
\mytrans{q_1}{\ep}{q_1\ q_2}
}
\end{minipage}
\hfill
\begin{minipage}{6.5cm} 
\tabrule{
(q_1,j,q_2,i)
}{
(q_1,j,q_3,i)
}{
\mytrans{q_1\ q_2}{\ep}{q_1\ q_3}
}
\end{minipage}
\hspace*{\fill}

\caption{Two additional inference rules for transitions of the form
$\mytrans{q_1}{\ep}{q_1\ q_2}$ and
$\mytrans{q_1\ q_2}{\ep}{q_1\ q_3}$.}
\label{NedSat:morerules}
\end{figure}
To extend the algorithm in
Figure~\ref{NedSat:generic_alg} to handle these additional
types of transitions requires more effort.
Up to now, all items $(q,j,q',i)$, with the exception of
$(\bot,0,\Xinit,0)$, were such that $j < i$.
If we had an item
$(q_1,j,q_2,i)$ in the agenda $\mynew$ and were
looking for items $(q',k,q_1,j)$ in $\mytable$, in order to
apply a transition $\mytrans{q_1\ q_2}{\ep}{q_3}$, then we
could be sure that we had access to all items $(q',k,q_1,j)$
that would ever be added to $\mytable$. This is because
$j<i$, and all items having $j$ as second input position had been
found at an earlier iteration of the algorithm.

However, if we add transitions of the form
$\mytrans{q_1}{\ep}{q_1\ q_2}$ and
$\mytrans{q_1\ q_2}{\ep}{q_1\ q_3}$,
we may obtain items of the form $(q,j,q',i)$ with $j=i$. It may then
happen that an item 
$(q',k,q_1,j)$ 
is added to $\mytable$ {\em after\/}
the item 
$(q_1,j,q_2,i)$ 
is taken from the agenda $\mynew$ and
processed. To avoid that we overlook any computation of the PDA,
we must change the algorithm to take into account that 
an item taken from the agenda may be of the form 
$(q',k,q_1,j)$, and we then need to find items of the form
$(q_1,j,q_2,i)$ already in the table,
with $j=i$, in order to apply a transition
$\mytrans{q_1\ q_2}{\ep}{q_3}$. 
We leave it to the reader to determine the precise changes this
requires to Figure~\ref{NedSat:generic_alg}, and to verify
that it is possible to implement these changes 
in such a way that the order of the time and space 
complexity remains unchanged.

\section{Earley's algorithm}
\label{NedSat:Earley}

In this section we will investigate
the top-down parsing strategy, and discuss tabulation
of the resulting PDAs.
Let us fix a CFG $\mygram = (\myterm, \mynont, S, \myrule)$
and let us assume that there is only one rule in $\myrule$
of the form $S \de \alpha$.
The stack symbols of the PDA that we will construct are
the so called {\em dotted rules\/}, defined as 
symbols of the form $A \de \alpha \bul \beta$ where
$A \de \alpha\beta$ is a rule from $\myrule$;
in words, a stack symbol is a rule in which a 
dot has been inserted somewhere in the right-hand side.
Intuitively, the dot separates the grammar symbols 
that have already been found to derive
substrings of the read input from those that
are still to be processed.
We will sometimes enclose dotted rules in
round brackets to enhance readability.

The alphabet of the PDA is the same as that of the CFG. 
The initial stack symbol is
$S \de \ \bul\alpha$, the final stack symbol is
$S \de \alpha\bul$, 
and the transitions are:
\begin{enumerate}
\item 
$\mytrans{(A \de \alpha \bul B \beta)}{\ep}{%
	(A \de \alpha \bul B \beta)\ (B \de\ \bul\gamma)}$
for all rules $A \de \alpha B \beta$ and
$B \de \gamma$;
\item
$\mytrans{(A \de \alpha \bul b \beta)}{b}{%
	(A \de \alpha b\bul \beta)}$
for each rule $A \de \alpha b \beta$, where
$b \in\myterm$;
\item 
$\mytrans{(A \de \alpha \bul B \beta)\ (B \de \gamma\bul)}{\ep}{%
        (A \de \alpha B \bul \beta)}$
for all rules $A \de \alpha B \beta$ and
$B \de \gamma$.
\end{enumerate}

Given a stack symbol $A \de \alpha \bul X \beta$, with
$X \in \myterm \cup \mynont$,  the indicated
occurrence of $X$ will here be called the {\em goal}.
The goal in the 
top-of-stack is the symbol that must be matched against the next
few unread input symbols.
Transitions of type~1 above predict rules with nonterminal $B$ in the
left-hand side, when $B$ is the goal
in the top-of-stack.
Transitions of type~2 move the dot over terminal goal $b$
in the top-of-stack, if that $b$ matches 
the next unread input symbol.
Finally, transitions of type~3 combine the top-most two
stack symbols, when the top-of-stack indicates
that the analysis of a rule
with $B$ in the left-hand side has been completed.
The current top-of-stack is removed, and in the new top-of-stack,
the dot is moved over the goal $B$.

Since the types of transition above are covered by what we discussed
in Section~\ref{NedSat:tabulation}, we may apply a subset of
the inference rules from Figures~\ref{NedSat:deduction}
and~\ref{NedSat:morerules} to obtain a tabular parsing
algorithm for the top-down strategy. This will result
in items of the form
$$(A \de \alpha \bul B \beta, j, B \de \gamma\bul\delta, i).$$
However, it can be easily verified that if there is
such an item in the table, and if some stack symbol
$A' \de \alpha' \bul B \beta'$ may occur on top of the stack
at position $j$, then at some point, the table will also contain
the item
$$(A' \de \alpha' \bul B \beta', j, B \de \gamma\bul\delta, i).$$
An implication of this is that the first component 
$A \de \alpha \bul B \beta$ of an
item represents redundant information, and may be removed
without affecting the correctness of the tabular algorithm.
(See~\cite[Section~1.2.2]{NE94b} for the exact conditions
that justify this simplification.)
After this simplification, we obtain the deduction system
in Figure~\ref{NedSat:earley}, which can be seen as
a specialized form of the tabular algorithm from the previous
section. It is also known as Earley's algorithm~\cite{EA70,AH72,GR76}.
Step~(\ref{NedSat:earley:init}) is called {\em initializer},
(\ref{NedSat:earley:pred}) is called {\em predictor},
(\ref{NedSat:earley:scan}) is called {\em scanner}, and
(\ref{NedSat:earley:comp}) is called {\em completer}.
\begin{figure}[t]
\begin{minipage}{7.0cm}
\tabrulelabel{NedSat:earley:init}{ \
}{
(0,S \de\ \bul\alpha,0)
}{
S \de\alpha
}
\tabrulelabel{NedSat:earley:pred}{
(j,A \de \alpha \bul B \beta,i)
}{
(i,B \de\ \bul\gamma,i)
}{
B \de \gamma
}
\end{minipage}
\hfill
\begin{minipage}{7.0cm}
\tabrulelabel{NedSat:earley:scan}{
(j,A \de \alpha \bul b \beta,i-1)
}{
(j,A \de \alpha b\bul \beta,i)
}{
b = a_i
}
\tabrulestartlabel{NedSat:earley:comp}{
(k,A \de \alpha \bul B \beta,j) \\
(j,B \de \gamma\bul,i)
}{
(k,A \de \alpha B \bul \beta,i)
}
\end{minipage}
\caption{Tabular top-down parsing, or Earley's algorithm.}
\label{NedSat:earley}
\end{figure}

As an example, consider the CFG with 
$\myterm=\{a, *, +\}$, $\mynont=\{S,E\}$ and with 
rules $S \de E$, $E \de E * E$, $E \de E + E$ 
and $E \de a$, and consider
the input string $w = a + a * a$. 
Now that items are 3-tuples, it is more convenient to
represent the table $\mytable$ as an upper triangular matrix
rather than a graph, as exemplified by Figure~\ref{NedSat:ex-earley}.
This matrix 
consists of sets $\mytable_{i,j}$, $i \leq j$, such that
$(A \de \alpha \bul \beta) \in \mytable_{i,j}$ if and only if
$(i, A \de \alpha \bul \beta, j) \in \mytable$.
The string $w$ is recognized since the final stack symbol 
$S \de E \bul$ is found in $\mytable_{0,5}$.  
Observe that $(0,S \de E \bul,5)$ can be 
derived from $(0,S \de\ \bul E,0)$ and
$(0,E \de E * E \bul,5)$ or
from $(0,S \de\ \bul E,0)$ and
$(0,E \de E + E \bul,5)$.
This indicates that $w$ is ambiguous.

\begin{figure}[t]
\begin{scriptsize}
$$
\begin{array}{r|@{}l@{}|@{}l@{}|@{}l@{}|@{}l@{}|@{}l@{}|@{}l@{}|}
\multicolumn{1}{l}{\ } &
\multicolumn{1}{l}{0} &
\multicolumn{1}{l}{1} &
\multicolumn{1}{l}{2} &
\multicolumn{1}{l}{3} &
\multicolumn{1}{l}{4} &
\multicolumn{1}{l}{5} \\
\cline{2-7} 
0 &
\begin{array}[t]{l}
S \de\ \bul E \\
E \de\ \bul E * E \\
E \de\ \bul E + E \\
E \de\ \bul a
\end{array} &
\begin{array}[t]{l}
E \de a \bul \\
S \de E \bul \\
E \de E \bul * E \\
E \de E \bul + E 
\end{array} &
\begin{array}[t]{l}
E \de E + \bul E 
\end{array} &
\begin{array}[t]{l}
E \de E + E \bul \\
S \de E \bul \\
E \de E \bul * E \\
E \de E \bul + E
\end{array} &
\begin{array}[t]{l}
E \de E * \bul E 
\end{array} &
\begin{array}[t]{l}
E \de E * E \bul \\
E \de E + E \bul \\
S \de E \bul \\
E \de E \bul * E \\
E \de E \bul + E
\end{array} 
\\
\cline{2-7}
\multicolumn{1}{l}{1} &
\multicolumn{1}{c|}{\ } &
\begin{array}[t]{l}
\\ \\ \\
\end{array} &
\begin{array}[t]{l}
\end{array} &
\begin{array}[t]{l}
\end{array} &
\begin{array}[t]{l}
\end{array} &
\begin{array}[t]{l}
\end{array}
\\
\cline{3-7}
\multicolumn{1}{l}{2} &
\multicolumn{2}{c|}{\ } &
\begin{array}[t]{l}
E \de\ \bul E * E \\
E \de\ \bul E + E \\
E \de\ \bul a
\end{array} &
\begin{array}[t]{l}
E \de a \bul \\
S \de E \bul \\
E \de E \bul * E \\
E \de E \bul + E 
\end{array} &
\begin{array}[t]{l}
E \de E * \bul E
\end{array} &
\begin{array}[t]{l}
E \de E * E \bul  \\
E \de E \bul * E \\
E \de E \bul + E 
\end{array}
\\
\cline{4-7}
\multicolumn{1}{l}{3} &
\multicolumn{3}{c|}{\ } &
\begin{array}[t]{l}
\\ \\ \\
\end{array} &
\begin{array}[t]{l}
\end{array} &
\begin{array}[t]{l}
\end{array}
\\
\cline{5-7}
\multicolumn{1}{l}{4} &
\multicolumn{4}{c|}{\ } &
\begin{array}[t]{l}
E \de\ \bul E * E \\
E \de\ \bul E + E \\
E \de\ \bul a
\end{array} &
\begin{array}[t]{l}
E \de a \bul  \\
E \de E \bul * E \\
E \de E \bul+ E 
\end{array}
\\
\cline{6-7}
\multicolumn{1}{l}{5} &
\multicolumn{5}{c|}{\ } &
\begin{array}[t]{l}
\\ \\ \\
\end{array} \\
\cline{7-7}
\end{array}
$$
\end{scriptsize}
\caption{Table $\mytable$ obtained by Earley's algorithm, represented as
upper triangular matrix.}
\label{NedSat:ex-earley}
\end{figure}

It can be easily verified that Earley's algorithm adds an item
$(j,A \de \alpha \bul \beta,i)$ to $\mytable$
if and only if:
\begin{enumerate}
\item $S \dem a_1 \cdots a_{j} A \gamma$, for some
$\gamma$, and 
\item $\alpha \dem a_{j+1} \cdots a_{i}$.
\end{enumerate}
In words, the existence of such an item in the table
means that there is a derivation from the
start symbol $S$ that reaches $A$, the part of that
derivation to the left of that occurrence of $A$ 
derives the input from position $0$ up to position $j$,
and the prefix $\alpha$ of the right-hand side of rule $A \de \alpha \beta$
derives the input from position $j$ up to position $i$.

The tabular algorithm of Figure~\ref{NedSat:earley}
runs in time $\order{\size{\mygram}^2 n^3}$ and space
$\order{\size{\mygram} n^2}$, for a CFG $\mygram$
and for an input string of length $n$.  
Both upper bounds can be easily derived from 
the general complexity results discussed
in Section~\ref{NedSat:tabulation}, taking into account
the simplification of items to 3-tuples.

To obtain a formulation of Earley's algorithm closer
to a practical implementation, such as that
in Figure~\ref{NedSat:generic_alg}, read the remarks
at the end of Section~\ref{NedSat:tabulation} concerning
the agenda and transitions that read the empty string.
Alternatively, one may also preprocess certain steps 
to avoid some of the problems with the
agenda during parse time, as discussed by~\cite{GR80}, who
also showed that the worst-case 
time complexity of Earley's algorithm can be improved to 
$\order{\size{\mygram} n^3}$.

\section{The Cocke-Kasami-Younger algorithm}
\label{NedSat:CKY}

Another parsing strategy is (pure) bottom-up parsing, 
which is also called shift-reduce parsing~\cite{SI88}.
It is particularly simple if the CFG 
$\mygram = (\myterm, \mynont, S, \myrule)$
is in Chomsky normal form, which means that each rule
is either of the form
$A\de a$, where $a\in\myterm$, or of the form
$A\de B\ C$, where $B,C\in\mynont$.
The set of stack symbols is the set of nonterminals
of the grammar, 
and the transitions are:
\begin{enumerate}
\item
$\mytrans{\ep}{a}{A}$ 
for each rule $A\de a$;
\item
$\mytrans{B\ C}{\ep}{A}$ for each rule $A\de B\ C$.
\end{enumerate}
A transition of type~1 consumes the next unread input symbol, and
pushes on the stack the nonterminal 
in the left-hand side of a corresponding rule.
A transition of type~2 can be applied if the top-most two 
stack symbols $B$ and $C$ are such that $B\ C$ is the right-hand side of
a rule, and it replaces $B$ and $C$ by the left-hand side $A$ of that rule.
Transitions of types~1 and~2 are called
{\em shift\/} and {\em reduce\/}, respectively; see also Section~\ref{NedSat:GLR}.
The final stack symbol is $S$. We deviate from the other sections in this
chapter however by 
assuming that the PDA starts with an empty stack, or
alternatively, that there is some imaginary initial stack symbol that is 
not in $\mynont$.

The transitions $\mytrans{B\ C}{\ep}{A}$
are of a type that
we have seen before, and 
in a tabular algorithm for the
PDA, such transitions
can be realized by the inference rule:
\begin{center}
\tabrule{
(k,B,j) \\
(j,C,i)
}{
(k,A,i)
}{
A \de B C
}
\end{center}
Here we use 3-tuples for items, since the first components
of the general 4-tuples are redundant, just as in the case
of Earley's algorithm above.
Transitions of the type
$\mytrans{\ep}{a}{A}$ are new, but they are similar to
transitions of the familiar form
$\mytrans{B}{a}{B\ A}$, where $B$ can be any stack symbol.
Because $B$ is irrelevant for deciding whether such a
transition can be applied, the expected inference rule
\begin{center}
\tabrule{
(j,B,i-1)
}{
(i-1,A,i)
}{
A\de a_i \\
B \in \mynont
}
\end{center}
can be simplified to
\begin{center}
\tabrule{
\ 
}{
(i-1,A,i)
}{
A\de a_i
}
\end{center}

A formulation of the tabular bottom-up
algorithm closer to a typical
implementation is given in Figure~\ref{NedSat:CKYalg}. 
This algorithm is also known as
the Cocke-Kasami-Younger (CKY) algorithm~\cite{YO67,AH72}.
Note that no agenda is needed.
It can be easily verified that the CKY algorithm adds an item
$(j,A,i)$ to $\mytable$
if and only if $A \dem a_{j+1} \cdots a_{i}$.
\begin{figure}[t]
\begin{enumerate}
\item Let $\mytable$ $=$ $\emptyset$.
\item For $i = 1, \ldots, n$ do:
\begin{enumerate}
\item For each rule $A \de a_i$, add $(i-1,A,i)$ to $\mytable$.
\item For $k = i-2, \ldots, 0$ and $j = k+1, \ldots, i-1$ do:
\begin{itemize}
\item
For each rule $A \de B\ C$ and all
$(k,B,j),(j,C,i)\in\mytable$,
add $(k,A,i)$ to $\mytable$.
\end{itemize}
\end{enumerate}
\item Recognize the input if $(0,S,n)\in\mytable$.
\end{enumerate}
\caption{Tabular bottom-up parsing, or the CKY algorithm.}
\label{NedSat:CKYalg}
\end{figure}

As an example, consider the CFG with 
$\myterm=\{a, b\}$, $\mynont=\{S,A\}$ 
and with rules $S \de SS$, $S \de AA$, $S \de b$,
$A \de AS$, $A \de AA$ and $A \de a$, and consider
the input string $w = aabb$.
The table $\mytable$ produced by the CKY algorithm 
is given in Figure~\ref{NedSat:ex-cky},
represented as an upper triangular matrix.
(Note that the sets $\mytable_{i,i}$, $0 \leq i \leq n$, 
on the diagonal of the matrix are always
empty and are therefore omitted.)
The string $w$ is recognized since the final stack symbol 
$S$ is found in $\mytable_{0,4}$.  

\begin{figure}[t]
\begin{small}
$$
\begin{array}{r|c|c|c|c|}
\multicolumn{1}{l}{\ } &
\multicolumn{1}{l}{1} &
\multicolumn{1}{l}{2} &
\multicolumn{1}{l}{3} &
\multicolumn{1}{l}{4} \\
\cline{2-5}
0 & 
A &
S, A &
S,A &
S,A
\\ \cline{2-5}
\multicolumn{1}{l}{1} &
\multicolumn{1}{c|}{\ } &
A &
A &
A
\\ \cline{3-5}
\multicolumn{1}{l}{2} &
\multicolumn{2}{c|}{\ } &
S &
S
\\ \cline{4-5}
\multicolumn{1}{l}{3} &
\multicolumn{3}{c|}{\ } &
S
\\ \cline{5-5}
\end{array}
$$
\end{small}
\caption{Table $\mytable$ obtained by the CKY algorithm.}
\label{NedSat:ex-cky}
\end{figure}

For a CFG $\mygram = (\myterm, \mynont, S, \myrule)$ in Chomsky normal
form and an input string of length $n$, the tabular algorithm of
Figure~\ref{NedSat:CKYalg} runs in time $\order{\size{R} n^3}$ and
space $\order{\size{\mynont} n^2}$.  
Again, these upper bounds can be easily derived from 
the general complexity results discussed
in Section~\ref{NedSat:tabulation}, taking into account
the simplification of items to 3-tuples.
Note that the CKY algorithm runs in time
proportional to the size of the grammar, since $\size{\mygram} = 
\order{\size{R}}$ for CFGs in Chomsky normal form.  
However,
known transformations to Chomsky normal form
may increase the size of the grammar by a square function~\cite{HA78}.

\section{Tabular LR parsing}
\label{NedSat:GLR}

A more complex parsing strategy is LR parsing~\cite{KN65,SI90}. 
Its main importance is that it results in
deterministic PDAs for many
practical CFGs for programming languages. 
For CFGs used in
natural language systems however, the resulting PDAs are 
typically nondeterministic. Although in this case
the advantages over simpler parsing strategies have
never been convincingly shown, the frequent treatment of 
nondeterministic LR parsing in recent literature
warrants its discussion here.

A distinctive feature of LR parsing is that commitment to a certain
rule is postponed until all grammar symbols in the right-hand side
of that rule have been found to generate appropriate substrings of the input. 
In particular, different rules for which this has not yet been accomplished
are processed simultaneously, without spending computational effort 
on any rule individually.
As in the case of Earley's algorithm, we need
dotted rules of the form $A\de\alpha\bul\beta$,
where the dot separates the 
grammar symbols in the right-hand side
that have already been found to derive
substrings of the read input
from those that are still to be processed. 
Whereas in the scanner step~(\ref{NedSat:earley:scan}) and
in the completer step~(\ref{NedSat:earley:comp}) from Earley's
algorithm (Figure~\ref{NedSat:earley}) each rule is 
individually processed by letting the dot traverse its right-hand side,
in LR parsing this traversal simultaneously affects 
{\em sets\/} of dotted rules.
Also the equivalent of the predictor step~(\ref{NedSat:earley:pred}) 
from Earley's algorithm is now an operation on sets of dotted rules. 
These operations are pre-compiled into stack symbols and 
transitions.

Let us fix a CFG $\mygram = (\myterm, \mynont, S, \myrule)$.
Assume $q$ is a set of dotted rules. We define
$\myclosure(q)$ as the smallest set of dotted rules such that:
\begin{enumerate}
\item $q \subseteq \myclosure(q)$, and
\item if $(A \de \alpha \bul B \beta) \in \myclosure(q)$ and
$(B \de \gamma)\in\myrule$, then $(B \de\ \bul \gamma)\in \myclosure(q)$.
\end{enumerate}
In words, we extend the set of dotted rules by those
that can be obtained by repeatedly applying 
an operation similar to the predictor step.
For a set $q$ of dotted rules and a grammar symbol $X\in\myterm\cup\mynont$,
we define:
\begin{eqnarray*}
\mygoto(q,X)&=& \myclosure(\{ (A \de \alpha X \bul \beta) \sep 
				(A \de \alpha \bul X \beta)\in q \})
\end{eqnarray*}
The manner in which the dot traverses through right-hand sides
can be related to the scanner step
of Earley's algorithm if $X\in\myterm$ or 
to the completer step if $X\in\mynont$.

The initial stack symbol $\Xinit$ is defined to be
$\myclosure(\{(S \de\ \bul\alpha) \sep (S \de \alpha)\in\myrule\})$; 
cf.\ the initializer step~(\ref{NedSat:earley:init})
of Earley's algorithm.
Other stack symbols
are those non-empty sets of dotted rules that can be derived 
from $\Xinit$ by means of repeated application of the 
goto function. More precisely,
$\mysym$ is the smallest set such that:
\begin{enumerate}
\item $\Xinit \in \mysym$, and
\item if $q\in \mysym$ and $\mygoto(q,X)=q'\neq\emptyset$ 
for some $X$, then $q' \in \mysym$.
\end{enumerate}
For technical reasons, we also need to add a special stack symbol
$\Xfinal$ to $\mysym$ that becomes the final stack symbol.
The transitions are:
\begin{enumerate}
\item $\mytrans{q_1}{a}{q_1\ q_2}$ for all $q_1,q_2\in\mysym$ and
each $a\in\myterm$ such that $\mygoto(q_1,a)=q_2$;
\item $\mytrans{q_0\ q_1\ \cdots\ q_m}{\ep}{q_0\ q'}$
for all $q_0,\ldots, q_m, q'\in\mysym$ and
each $(A \de \alpha \bul) \in q_m$ such that $\size{\alpha}=m$
and $q' = \mygoto(q_0,A)$;
\item $\mytrans{q_0\ q_1\ \cdots\ q_m}{\ep}{\Xfinal}$
for all $q_0,\ldots, q_m\in\mysym$ and each
$(S \de \alpha \bul) \in q_m$ such that $\size{\alpha}=m$
and $q_0 = \Xinit$.
\end{enumerate}
The first type of transition is called {\em shift\/}.
It can be seen as the pre-compilation of the scanner step followed
by repeated application of the predictor step.
Note that only one transition is applied
for each input symbol that is read, 
independent of the number of dotted rules
in the sets $q_1$ and $q_2$.
The second type of transition is called {\em reduction}.
It can be applied when the symbol on top of the stack
contains a dotted rule with the dot at the end of the right-hand side.
First, 
as many symbols are popped from the stack as that right-hand side
is long,
and then a symbol $q'=\mygoto(q_0,A)$ is pushed on the stack. This is related
to the completer step from Earley's algorithm.
The third type of transition is very similar to the second.
It is only applied once, when the start symbol has been found to
generate (a prefix of) the input.

For tabular LR parsing, we apply the same framework
as in the previous sections, to obtain
Figure~\ref{NedSat:LR}. A slight difficulty is caused 
by the new types of transition
$\mytrans{q_0\ \cdots\ q_m}{\ep}{q_0\ q'}$ and
$\mytrans{q_0\ \cdots\ q_m}{\ep}{\Xfinal}$,
but these can be handled by a straightforward generalization
of inference rules from Figures~\ref{NedSat:deduction} and~\ref{NedSat:morerules}.
Note that we need 4-tuple items here rather than 3-tuple items
as in the previous two sections. 
\begin{figure}[t]
\begin{minipage}[t]{7.5cm}
\tabrulestartlabel{NedSat:LR:init}{ \
}{
(\bot,0,\Xinit,0)
}

\tabrulelabel{NedSat:LR:shift}{
(q',j,q_1,i-1)
}{
(q_1,i-1,q_2,i)
}{
\mygoto(q_1,a_i)=q_2\!\!\!\!\!\!\!
}
\end{minipage}
\hspace{2ex}
\begin{minipage}[t]{7.5cm}
\tabrulelabel{NedSat:LR:reduce}{
(q_0,j_0,q_1,j_1) \\
(q_1,j_1,q_2,j_2) \\[-0.9ex]
\vdots \\[-0.9ex]
(q_{m-1},j_{m-1},q_m,j_m) \\
}{
(q_0,j_0,q',j_m)
}{
(A \de \alpha \bul) \in q_m \!\!\! \\
\size{\alpha}=m \\
q' = \mygoto(q_0,A) \!\!\!
}

\tabrulelabel{NedSat:LR:accept}{
(\bot,0,q_0,j_0) \\
(q_0,j_0,q_1,j_1) \\
(q_1,j_1,q_2,j_2) \\[-0.9ex]
\vdots \\[-0.9ex]
(q_{m-1},j_{m-1},q_m,j_m) \!\!\!\!\! \\
}{
(\bot,0,\Xfinal,j_m)
}{
(S \de \alpha \bul) \in q_m \\
\size{\alpha}=m \\
q_0 = \Xinit
}
\end{minipage}
\caption{Tabular LR parsing, or generalized LR parsing.}
\label{NedSat:LR}
\end{figure}
Tabular LR parsing is also known as 
{\em generalized\/} LR parsing~\cite{TO86,TO87}. In the literature
on generalized LR parsing, but only there, the table
$\mytable$ of items is often called a {\em graph-structured stack}.

\begin{figure}[t]
\begin{center}
\includegraphics{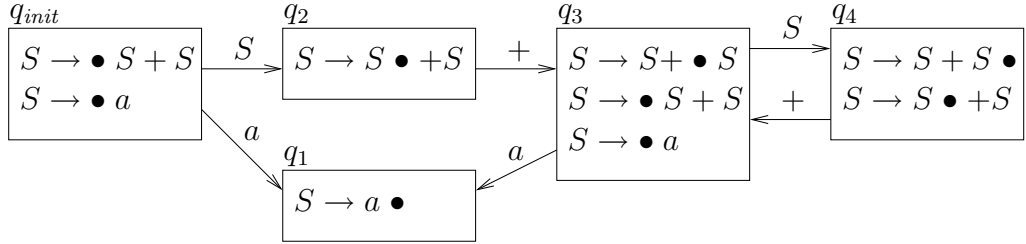}
\end{center}
\caption{The set of stack symbols, excluding $\Xfinal$, 
and the $\mygoto$ function.}
\label{NedSat:LRaut}
\end{figure}

\begin{figure}[t]
\begin{center}
\includegraphics{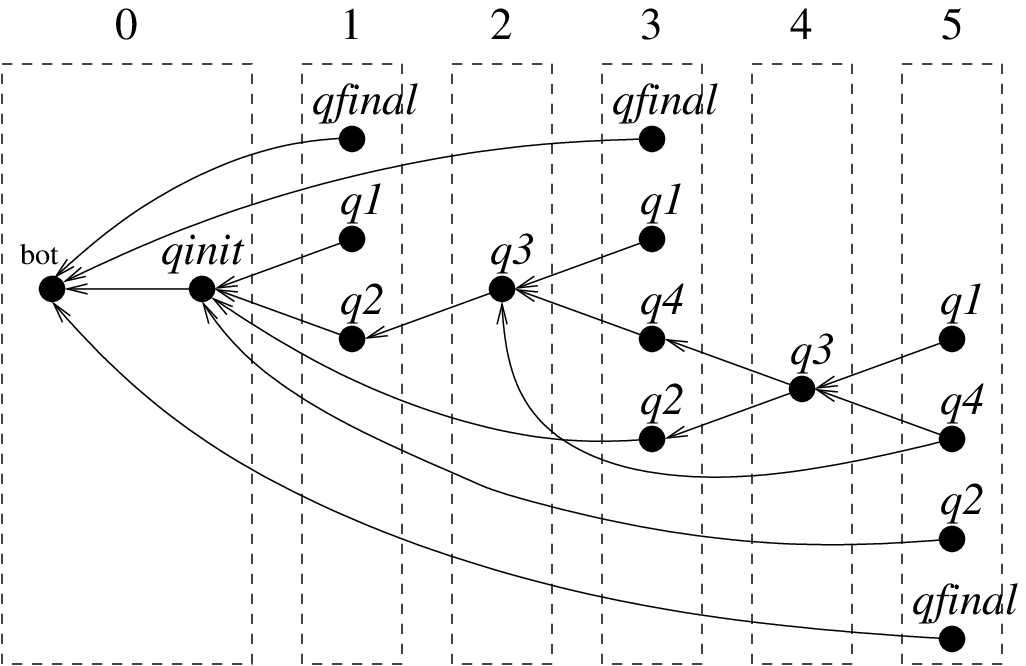}
\end{center}
\caption{Table $\mytable$ obtained by tabular LR parsing.}
\label{NedSat:LRgraph}
\end{figure}

As an example, consider the grammar with the 
rules $S\de S+S$ and $S\de a$. Apart from
$\Xfinal$, the stack symbols of the PDA are
represented 
in Figure~\ref{NedSat:LRaut} as rectangles enclosing
sets of dotted rules. There is an
arc from stack symbol $q$ to stack symbol $q'$ labelled
by $X$ to denote that $\mygoto(q,X)=q'$.
For the input $a+a+a$, 
the table $\mytable$ is given by
Figure~\ref{NedSat:LRgraph}. 
Note that $q_3$ at position
4 has two outgoing arcs, since it can arise by a shift with
$+$ from $q_4$ or from $q_2$. Also note that
$(\bot,0,\Xfinal,5)$ is found twice, once from 
$(\bot,0,\Xinit,0)$,
$(\Xinit,0,q_2,1)$,
$(q_2,1,q_3,2)$,
$(q_3,2,q_4,5)$,
and once from
$(\bot,0,\Xinit,0)$,
$(\Xinit,0,q_2,3)$,
$(q_2,3,q_3,4)$,
$(q_3,4,q_4,5)$, in both cases
by means of $(S\de S+S\bul)\in q_4$, with
$\size{S+S}=3$.
This indicates that the input $a+a+a$ is ambiguous.

If the grammar at hand does not contain 
rules of the form $A\de\ep$, then the tabular
algorithm from Figure~\ref{NedSat:LR} can be reformulated
in a way very similar to the algorithm
from Figure~\ref{NedSat:generic_alg}. If there are
rules of the form $A\de\ep$ however, the handling of
the agenda is complicated, due to problems similar to those 
we discussed at the end of Section~\ref{NedSat:tabulation}. 
This issue is investigated by~\cite{NO91,NE96e}.

We now analyze the time and space complexity of tabular LR parsing.
Let us fix a CFG 
$\mygram = (\myterm, \mynont, S, \myrule)$. 
Let $p$ be the length of the longest 
right-hand side of a rule in $R$ and
let $n$ be the length of the input string.  Once again, we assume that
$\mytable$ is implemented  as a square array of size $n+1$. 
Consider the reduction step~(\ref{NedSat:LR:reduce}) in Figure~\ref{NedSat:LR}.
Each application of this step is uniquely
identified by $m+1\leq p+1$ input positions and $\size{Q}\size{R}$ combinations of
stack symbols. 
The expression $\size{Q}\size{R}$ is due to the fact that, once a stack
symbol $q_0$ and a rule $A \de X_1 X_2 \cdots X_{m}$ have been selected
such that $(A \de \; \bul \! X_1 X_2 \cdots X_{m}) \in q_0$, 
then stack symbols $q_i$, $1 \leq i \leq m$, and $q'$ are uniquely 
determined by $q_1 = \mygoto(q_0, X_1)$, 
$q_2 = \mygoto(q_1, X_2)$, $\ldots$, $q_{m} = \mygoto(q_{m-1}, X_{m})$
and $q' =  \mygoto(q_0, A)$. 
(As can be easily verified, a derivable stack of which the top-most symbol 
$q_m$ contains $(A \de X_1 X_2 \cdots X_{m}\bul)$ must necessarily have top-most
symbols $q_0q_1\cdots q_m$ with the above constraints.)
Since a single application of this step can easily be carried
out in time $\order{p}$, we conclude 
the total amount of time required
by all applications of the step is $\order{\size{Q}\size{R}p n^{p+1}}$.
This is also the worst-case time complexity of the algorithm, 
since the running time is dominated by the reduction step~(\ref{NedSat:LR:reduce}).
From the general complexity results discussed
in Section~\ref{NedSat:tabulation} it follows that
the worst-case space complexity is $\order{\size{Q}^2 n^2}$. 

We observe that while the above time bound is polynomial
in the length of the input string, it can be much worse than the corresponding
bounds for Earley's algorithm or for the CKY algorithm, since $p$
is not bounded.
A solution to this problem has been discussed by~\cite{KI91,NE96} and consists in 
splitting each reduction into $\order{p}$ transitions
of the form $\mytrans{q' q''}{\ep}{q}$. In this way,
the maximum length of transitions becomes independent of the grammar.
This results in 
tabular implementations of LR parsing with cubic time complexity in the
length of the input.
We furthermore observe that the term $\size{Q}$ in the above 
bounds depends on the specific structure of $\mygram$, and may
grow exponentially with $\size{\mygram}$~\cite[Proposition~6.46]{SI90}.

\section{Parse trees}
\label{NedSat:forests}

As stated in Section~\ref{NedSat:intro}, recognition is the process
of determining whether an input string is in the language
described by a grammar or automaton, and 
parsing is the process of determining the parse trees of an input 
string according to a grammar.   Although the algorithms
we have discussed up to now are recognition algorithms, they
can be easily extended to become parsing algorithms, as we show
in this section.  In what follows we assume a fixed CFG
$\mygram = (\myterm, \mynont, S, \myrule)$
and an input string $w = a_1 \cdots a_n \in \myterm^\ast$.

Since the number of parse trees can be exponential in the length
of the input string, and even infinite when $\mygram$ is cyclic,
one first needs to find a way to compactly represent the set of all 
parse trees.   This is usually done through a CFG $\mygram_w$, called 
{\em parse forest\/}, defined as follows.  The alphabet of 
$\mygram_w$ is the same as that of $\mygram$, and the nonterminals of
$\mygram_w$ have the form $(j,A,i)$, where $A \in \mynont$ and 
$0 \leq j \leq i \leq n$.  The start symbol of $\mygram_w$ is $(0,S,n)$.
The rules of $\mygram_w$ include at least those of the form
$(i_0,A,i_m)\de (i_0,X_1,i_1) \cdots (i_{m-1},X_m,i_m)$,
where (i)~$(A \de X_1 \cdots X_m)\in\myrule$, 
(ii)~$S \dem a_1 \cdots a_{i_0} A a_{i_m+1} \cdots a_n$,
and (iii)~$X_j \dem a_{i_{j-1}+1} \cdots a_{i_j}$ for $1 \leq j \leq m$,
and those of the form $(i-1,a_i,i)\de a_i$.
However, $\mygram_w$ may also contain rules 
$(i_0,A,i_m)\de (i_0,X_1,i_1) \cdots (i_{m-1},X_m,i_m)$
that violate constraints~(ii) or~(iii) above.
Such rules
cannot be part of any derivation of a terminal string from $(0,S,n)$
and can be eliminated by a process that is called
{\em reduction}. Reduction can be carried out in linear 
time in the size of $\mygram_w$~\cite{SI88}.

It is not difficult to show that the 
parse forest $\mygram_w$ generates a finite language, which is either
$\{w\}$ if $w$ is in the language generated by $\mygram$, or $\emptyset$
otherwise.  Furthermore, there is a one-to-one correspondence between
parse trees according to $\mygram_w$ and parse trees of $w$
according to $\mygram$, with corresponding parse trees being isomorphic.

To give a concrete example, let us consider the CKY algorithm
presented in Section~\ref{NedSat:CKY}. 
In order to extend this recognition algorithm to a parsing algorithm, 
we may construct the parse forest $\mygram_w$ with 
rules of the form $(j,A,i) \de (j,B,k)\ (k,C,i)$, 
where $(A \de B\ C)\in\myrule$ and $(j,B,k),(k,C,i)\in\mytable$,
rules of the form $(i-1,A,i)\de (i-1,a_i,i)$, where $(A \de a_i)\in\myrule$, 
and rules of the form $(i-1,a_i,i)\de a_i$.
Such rules can be constructed during the computation of the table $\mytable$.
In order to perform reduction on $\mygram_w$, one may visit
the nonterminals of $\mygram_w$ starting from $(0,S,n)$, 
following the rules in a top-down fashion,
eliminating 
the nonterminals and the associated rules that are never reached.
From the resulting parse forest $\mygram_w$, individual parse trees can 
be extracted in time proportional to the size of the parse tree
itself, which in the case of CFGs in Chomsky normal form is $\order{n}$.
One may also extract parse trees directly from table $\mytable$, but
the time complexity then becomes $\order{\size{\mygram} n^2}$~\cite{AH72,GR76}.

Consider the table $\mytable$ from Figure~\ref{NedSat:ex-cky}, 
which was produced by the CKY algorithm with $w = aabb$
and $\mygram = (\myterm, \mynont, S, \myrule)$, 
where $\myterm=\{a, b\}$, $\mynont=\{S,A\}$ and
$\myrule = \{ S \de SS, S \de AA, S \de b, A \de AS, A \de AA, A \de a \}$.  
The method presented above constructs the parse forest
$\mygram_w = (\myterm, \mynont_w, (0, S, 4), \myrule_w)$, 
where $\mynont_w \subseteq \{ (j, B, i) \sep B \in \mynont, \; 0 \leq j < i \leq 4 \}$ 
and $\myrule_w$ contains the rules in Figure~\ref{NedSat:ex-pforest}.
Rules that are eliminated by reduction are marked by $\dag$.

\begin{figure}[t]
\hspace*{\fill}
\begin{minipage}[t]{6cm}
$$
\begin{array}[t]{rcll}
(0, a, 1) & \de & a                   \\
(1, a, 2) & \de & a                   \\
(2, b, 3) & \de & b                   \\
(3, b, 4) & \de & b 			\\
(0, A, 1) & \de & (0, a, 1)           \\
(1, A, 2) & \de & (1, a, 2)           \\
(2, S, 3) & \de & (2, b, 3)           \\
(3, S, 4) & \de & (3, b, 4)           \\[.5ex]

(0, S, 2) & \de & (0, A, 1)\ (1, A, 2) \\[.5ex]

(0, A, 2) & \de & (0, A, 1)\ (1, A, 2)\ \dag \\[.5ex]

(1, A, 3) & \de & (1, A, 2)\ (2, S, 3) \\[.5ex]

(2, S, 4) & \de & (2, S, 3)\ (3, S, 4) 
\end{array}
$$
\end{minipage}
\hfill
\begin{minipage}[t]{6cm}
$$
\begin{array}[t]{rcll}
(0, S, 3) & \de & (0, A, 1)\ (1, A, 3) \\
(0, S, 3) & \de & (0, S, 2)\ (2, S, 3) \\[.5ex]

(0, A, 3) & \de & (0, A, 1)\ (1, A, 3)\ \dag \\
(0, A, 3) & \de & (0, A, 2)\ (2, S, 3)\ \dag \\[.5ex]

(1, A, 4) & \de & (1, A, 2)\ (2, S, 4) \\
(1, A, 4) & \de & (1, A, 3)\ (3, S, 4) \\[.5ex]

(0, S, 4) & \de & (0, A, 1)\ (1, A, 4) \\
(0, S, 4) & \de & (0, S, 2)\ (2, S, 4) \\
(0, S, 4) & \de & (0, S, 3)\ (3, S, 4) \\[.5ex]

(0, A, 4) & \de & (0, A, 1)\ (1, A, 4)\ \dag \\
(0, A, 4) & \de & (0, A, 2)\ (2, S, 4)\ \dag \\
(0, A, 4) & \de & (0, A, 3)\ (3, S, 4)\ \dag 
\end{array}
$$
\end{minipage}
\hspace*{\fill}
\caption{Parse forest associated with table $\mytable$ from
Figure~\ref{NedSat:ex-cky}.}
\label{NedSat:ex-pforest}
\end{figure}

If $\mygram$ is in Chomsky normal form, then
we have $\size{\mygram_w} = \order{\size{\mygram} n^{3}}$.
For general CFGs, however, we have 
$\size{\mygram_w} = \order{\size{\mygram} n^{p+1}}$, 
where $p$ is the length of the longest 
right-hand side of a rule in $\mygram$. 
In practical parsing applications this higher space complexity 
is usually avoided by applying 
the following method, which is based on~\cite{LA74,BI89}. 
In place of computing $\mygram_w$, one constructs an alternative
CFG containing rules of the form
$t \de t_1 \cdots t_m$, where $t,t_1,\ldots,t_m\in\mytable$
such that item $t$ was derived from items $t_1,\ldots,t_m$
via an inference rule with $m$ antecedents. 
Parse trees according to this new CFG can be extracted as 
usual. 
From these trees, the desired parse trees for $w$ according 
to $\mygram$ can be easily obtained by elementary tree editing
operations such as node relabelling and node erasing.   
The precise editing algorithm that 
should be applied depends on the deduction 
system underlying the adopted recognition algorithm.

If the adopted recognition algorithm has inference
rules with no more than $m=2$ antecedents,
then the space complexity of the parsing method discussed above,
expressed as a function of the length $n$ of the input string, 
is $\order{n^{3}}$.  Note that $m = 2$ in the case of Earley's
algorithm, and this also holds in practical implementations of 
tabular LR parsing, as discussed at the end of 
Section~\ref{NedSat:GLR}. 
The space complexity in the
size of $\mygram$ may be larger than $\order{\size{\mygram}}$, however;
it is $\order{\size{\mygram}^2}$ in the case of 
Earley's algorithm and even exponential in the case of
tabular LR parsing.

The parse forest representation is originally due 
to~\cite{BA64}, with states of a finite automaton in place of 
positions in an input string.
Parse forests have also been discussed
by~\cite{CO70,SH76,TO86,LE90}.
Similar ideas were proposed for tree-adjoining grammars
by~\cite{VI93,LA92}.

\section{Further references}
\label{NedSat:literature}

In this chapter we have restricted ourselves to tabulation for
context-free parsing, on the basis of PDAs.
A similar kind of tabulation was also developed for
tree-adjoining grammars
on the basis of an extended type of PDA~\cite{AL00a}. 
Tabulation for an even more general type of PDA
was discussed by~\cite{BA96}.

A further restriction we have made is that the input
to the parser must be a string. 
Context-free parsing can however be generalized to input consisting
of a finite automaton.  
Finite automata without cycles used in speech recognition systems are
also referred to as word graphs or word lattices \cite{AU95}.
The parsing methods developed in this chapter can be 
easily adapted to parsing of finite automata, by 
manipulating states of an input automaton in 
place of positions in an input string.
This technique can be traced back to~\cite{BA64},
which we mentioned before in Section~\ref{NedSat:forests}.

PDAs are usually considered to read
input from left to right, and the forms
of tabulation that we discussed follow that directionality.%
\footnote{There are alternative forms of tabulation that do 
not adopt the left-to-right mode of processing from
the PDA~\cite{AH68,NE96c}.}
For types of tabular parsing that are not strictly
in one direction, such as head-driven 
parsing~\cite{SI97} and island-driven parsing~\cite{SA94},
it is less appealing to take PDAs
as starting point.

Earley's algorithm and the CKY algorithm
run in cubic time in the length of the input string. 
An asymptotically faster method for context-free parsing
has been developed by~\cite{VA75}, using a reduction 
from context-free recognition to Boolean matrix multiplication. 
An inverse reduction from Boolean matrix multiplication
to context-free recognition has been presented by~\cite{LE01}, 
providing evidence that asymptotically faster methods for 
context-free recognition might not be of practical interest. 

The extension of tabular parsing with weights or
probabilities has been considered 
by~\cite{LY74} for Earley's algorithm,
by~\cite{TE73} for the CKY algorithm,
and by~\cite{LA93} for tabular LR parsing.
Deduction systems for parsing extended with weights are discussed by~\cite{GO99}.

\bibliographystyle{plain}
\bibliography{/home/markjan/bib/refs}

\newcommand{\noop}[1]{}\newcommand{\id}[1]{#1}
\begin{thebibliography}{10}

\bibitem{AH68}
A.V. Aho, J.E. Hopcroft, and J.D. Ullman.
\newblock Time and tape complexity of pushdown automaton languages.
\newblock {\em Information and Control}, 13:186--206, 1968.

\bibitem{AH72}
A.V. Aho and J.D. Ullman.
\newblock {\em Parsing}, volume~1 of {\em The Theory of Parsing, Translation
  and Compiling}.
\newblock Prentice-Hall, 1972.

\bibitem{AL00a}
M.~A. Alonso~Pardo, M.-J. Nederhof, and E.~Villemonte de~la Clergerie.
\newblock Tabulation of automata for tree-adjoining languages.
\newblock {\em Grammars}, 3:89--110, 2000.

\bibitem{AU95}
H.~Aust, M.~Oerder, F.~Seide, and V.~Steinbiss.
\newblock The {P}hilips automatic train timetable information system.
\newblock {\em Speech Communication}, 17:249--262, 1995.

\bibitem{BA64}
Y.~Bar-Hillel, M.~Perles, and E.~Shamir.
\newblock On formal properties of simple phrase structure grammars.
\newblock In Y.~Bar-Hillel, editor, {\em Language and Information: Selected
  Essays on their Theory and Application}, chapter~9, pages 116--150.
  Addison-Wesley, 1964.

\bibitem{BI89}
S.~Billot and B.~Lang.
\newblock The structure of shared forests in ambiguous parsing.
\newblock In {\em 27th Annual Meeting of the Association for Computational
  Linguistics, Proceedings of the Conference}, pages 143--151, Vancouver,
  British Columbia, Canada, June 1989.

\bibitem{CO70}
J.~Cocke and J.T. Schwartz.
\newblock {\em Programming Languages and Their Compilers --- Preliminary
  Notes}, pages 184--206.
\newblock Courant Institute of Mathematical Sciences, New York University,
  second revised version, April 1970.

\bibitem{CO70a}
S.A. Cook.
\newblock Path systems and language recognition.
\newblock In {\em ACM Symposium on Theory of Computing}, pages 70--72, 1970.

\bibitem{EA70}
J.~Earley.
\newblock An efficient context-free parsing algorithm.
\newblock {\em Communications of the ACM}, 13(2):94--102, February 1970.

\bibitem{GO99}
J.~Goodman.
\newblock Semiring parsing.
\newblock {\em Computational Linguistics}, 25(4):573--605, 1999.

\bibitem{GR76}
S.L. Graham and M.A. Harrison.
\newblock Parsing of general context free languages.
\newblock In {\em Advances in Computers}, volume~14, pages 77--185. Academic
  Press, New York, NY, 1976.

\bibitem{GR80}
S.L. Graham, M.A. Harrison, and W.L. Ruzzo.
\newblock An improved context-free recognizer.
\newblock {\em ACM Transactions on Programming Languages and Systems},
  2(3):415--462, July 1980.

\bibitem{HA78}
M.A. Harrison.
\newblock {\em Introduction to Formal Language Theory}.
\newblock Addison-Wesley, 1978.

\bibitem{KI91}
J.R. Kipps.
\newblock {GLR} parsing in time {${\cal O}(n^3)$}.
\newblock In M.~Tomita, editor, {\em Generalized {LR} Parsing}, chapter~4,
  pages 43--59. Kluwer Academic Publishers, 1991.

\bibitem{KN65}
D.E. Knuth.
\newblock On the translation of languages from left to right.
\newblock {\em Information and Control}, 8:607--639, 1965.

\bibitem{LA74}
B.~Lang.
\newblock Deterministic techniques for efficient non-deterministic parsers.
\newblock In {\em Automata, Languages and Programming, 2nd Colloquium},
  volume~14 of {\em Lecture Notes in Computer Science}, pages 255--269,
  Saarbr{\"u}cken, 1974. Springer-Verlag.

\bibitem{LA92}
B.~Lang.
\newblock Recognition can be harder than parsing.
\newblock {\em Computational Intelligence}, 10(4):486--494, 1994.

\bibitem{LA93}
A.~Lavie and M.~Tomita.
\newblock {GLR$^*$} -- an efficient noise-skipping parsing algorithm for
  context free grammars.
\newblock In {\em Third International Workshop on Parsing Technologies}, pages
  123--134, Tilburg (The Netherlands) and Durbuy (Belgium), August 1993.

\bibitem{LE01}
L.~Lee.
\newblock Fast context-free grammar parsing requires fast boolean matrix
  multiplication.
\newblock {\em Journal of the ACM}, 49(1):1--15, 2001.

\bibitem{LE93}
R.~Leermakers.
\newblock {\em The Functional Treatment of Parsing}.
\newblock Kluwer Academic Publishers, 1993.

\bibitem{LE90}
H.~Leiss.
\newblock On {K}ilbury's modification of {E}arley's algorithm.
\newblock {\em ACM Transactions on Programming Languages and Systems},
  12(4):610--640, October 1990.

\bibitem{LY74}
G.~Lyon.
\newblock Syntax-directed least-errors analysis for context-free languages: A
  practical approach.
\newblock {\em Communications of the ACM}, 17(1):3--14, January 1974.

\bibitem{NE96c}
M.-J. Nederhof.
\newblock Reversible pushdown automata and bidirectional parsing.
\newblock In J.~Dassow, G.~Rozenberg, and A.~Salomaa, editors, {\em
  Developments in Language Theory II}, pages 472--481. World Scientific,
  Singapore, 1996.

\bibitem{NE96e}
M.-J. Nederhof and J.J. Sarbo.
\newblock Increasing the applicability of {LR} parsing.
\newblock In H.~Bunt and M.~Tomita, editors, {\em Recent Advances in Parsing
  Technology}, chapter~3, pages 35--57. Kluwer Academic Publishers, 1996.

\bibitem{NE96}
M.-J. Nederhof and G.~Satta.
\newblock Efficient tabular {LR} parsing.
\newblock In {\em 34th Annual Meeting of the Association for Computational
  Linguistics, Proceedings of the Conference}, pages 239--246, Santa Cruz,
  California, USA, June 1996.

\bibitem{NE94b}
M.J. Nederhof.
\newblock {\em Linguistic Parsing and Program Transformations}.
\newblock PhD thesis, University of Nijmegen, 1994.

\bibitem{NO91}
R.~Nozohoor-Farshi.
\newblock {GLR} parsing for $\varepsilon$-grammars.
\newblock In M.~Tomita, editor, {\em Generalized {LR} Parsing}, chapter~5,
  pages 61--75. Kluwer Academic Publishers, 1991.

\bibitem{SA94}
G.~Satta and O.~Stock.
\newblock Bidirectional context-free grammar parsing for natural language
  processing.
\newblock {\em Artificial Intelligence}, 69:123--164, 1994.

\bibitem{SH76}
B.A. Sheil.
\newblock Observations on context-free parsing.
\newblock {\em Statistical Methods in Linguistics}, pages 71--109, 1976.

\bibitem{SH95}
S.M. Shieber, Y.~Schabes, and F.C.N. Pereira.
\newblock Principles and implementation of deductive parsing.
\newblock {\em Journal of Logic Programming}, 24:3--36, 1995.

\bibitem{SI97}
K.~Sikkel.
\newblock {\em Parsing Schemata}.
\newblock Springer-Verlag, 1997.

\bibitem{SI88}
S.~Sippu and E.~Soisalon-Soininen.
\newblock {\em Parsing Theory, Vol. I: Languages and Parsing}, volume~15 of
  {\em EATCS Monographs on Theoretical Computer Science}.
\newblock Springer-Verlag, 1988.

\bibitem{SI90}
S.~Sippu and E.~Soisalon-Soininen.
\newblock {\em Parsing Theory, Vol. II: LR($k$) and LL($k$) Parsing}, volume~20
  of {\em EATCS Monographs on Theoretical Computer Science}.
\newblock Springer-Verlag, 1990.

\bibitem{TE73}
R.~Teitelbaum.
\newblock Context-free error analysis by evaluation of algebraic power series.
\newblock In {\em Conference Record of the Fifth Annual ACM Symposium on Theory
  of Computing}, pages 196--199, 1973.

\bibitem{TH84}
H.~Thompson and G.~Ritchie.
\newblock Implementing natural language parsers.
\newblock In T.~O'Shea and M.~Eisenstadt, editors, {\em Artificial
  Intelligence: Tools, Techniques, and Applications}, chapter~9, pages
  245--300. Harper \& Row, New York, 1984.

\bibitem{TO86}
M.~Tomita.
\newblock {\em Efficient Parsing for Natural Language}.
\newblock Kluwer Academic Publishers, 1986.

\bibitem{TO87}
M.~Tomita.
\newblock An efficient augmented-context-free parsing algorithm.
\newblock {\em Computational Linguistics}, 13:31--46, 1987.

\bibitem{VA75}
L.G. Valiant.
\newblock General context-free recognition in less than cubic time.
\newblock {\em Journal of Computer and System Sciences}, 10:308--315, 1975.

\bibitem{VI93}
K.~Vijay-Shanker and D.J. Weir.
\newblock The use of shared forests in tree adjoining grammar parsing.
\newblock In {\em Sixth Conference of the European Chapter of the Association
  for Computational Linguistics, Proceedings of the Conference}, pages
  384--393, Utrecht, The Netherlands, April 1993.

\bibitem{BA96}
E.~Villemonte de~la Clergerie and F.~Barth{\'e}lemy.
\newblock Information flow in tabular interpretations for generalized push-down
  automata.
\newblock {\em Theoretical Computer Science}, 199:167--198, 1998.

\bibitem{YO67}
D.H. Younger.
\newblock Recognition and parsing of context-free languages in time $n^3$.
\newblock {\em Information and Control}, 10:189--208, 1967.

\end{thebibliography}

\end{document}